\documentclass[letterpaper]{article}
\usepackage{uai2020}

\usepackage{times}

\usepackage[numbers, compress]{natbib}

\usepackage[margin=1in]{geometry}
\usepackage[utf8]{inputenc} 
\usepackage[T1]{fontenc}    
\usepackage{hyperref}       
\usepackage{url}            
\usepackage{booktabs}       
\usepackage{amsfonts}       
\usepackage{nicefrac}       
\usepackage{microtype}      
\usepackage{subfigure}
\usepackage{amsmath}
\usepackage{amssymb}
\usepackage{adjustbox}
\usepackage{multirow}
\usepackage{algorithm}
\usepackage[noend]{algpseudocode}
\usepackage{accents}
\usepackage{epstopdf}
\usepackage{hyperref}

\newtheorem{theorem}{Theorem}[section]

\newtheorem{definition}{Definition}[section]

\DeclareMathOperator*{\argmax}{arg\,max}
\newcommand{\R}{{\mathbb R}} 
\newcommand{\bbS}{{\mathbb S}} 
\newcommand{\E}{{\mathbb E}} 
 
\newcommand{\N}{{\mathcal N}} 
\newcommand{\A}{{\mathbb A}}

\DeclareMathOperator{\bx}{\mathbf{x}}
\DeclareMathOperator{\GP}{\mathcal{GP}}
\newcommand{\EI}{\text{EI}}

\newcommand{\by}{{\mathbf{y}}}
\newcommand{\bk}{{\mathbf{k}}}

\newcommand{\D}{{\mathcal{D}}}
\newcommand{\mI}{\mathsf{I}}
\newcommand{\mK}{\mathsf{K}}

\newcommand{\figref}[2][]{{Figure~\ref{#2}#1}}

\newcommand{\Var}{\text{Var}}
\newcommand{\Cov}{\text{Cov}}
\newcommand{\bg}{\textbf{g}}

\title{Efficient Rollout Strategies for Bayesian Optimization}

\author{
    Eric Hans Lee \\
    Cornell University, CS \\
    \texttt{erichanslee@cs.cornell.edu} \\
    \And
    David Eriksson \\
    Uber AI \\
    \texttt{eriksson@uber.com} \\
    \And
    Bolong Cheng \\
    SigOpt \\
    \texttt{harvey@sigopt.com} \\
    \AND 
    Michael McCourt \\
    SigOpt \\
    \texttt{mike@sigopt.com} \\
    \And
    David Bindel \\
    Cornell University, CS \\
    \texttt{bindel@cs.cornell.edu } \\
}

\begin{document}
\maketitle
\graphicspath{{.}}

\begin{abstract}
Bayesian optimization (BO) is a class of sample-efficient global optimization methods, where a probabilistic model conditioned on previous observations is used to determine future evaluations via the optimization of an acquisition function. 
Most acquisition functions are myopic, meaning that they only consider the impact of the next function evaluation. 
Non-myopic acquisition functions consider the impact of the next $h$ function evaluations and are typically computed through rollout, in which $h$ steps of BO are simulated. 
These rollout acquisition functions are defined as $h$-dimensional integrals, and are expensive to compute and optimize. 
We show that a combination of quasi-Monte Carlo, common random numbers, and control variates significantly reduce the computational burden of rollout.
We then formulate a policy-search based approach that removes the need to optimize the rollout acquisition function. 
Finally, we discuss the qualitative behavior of rollout policies in the setting of multi-modal objectives and model error.

\end{abstract}

\section{INTRODUCTION}
\label{sec:introduction}
Bayesian optimization (BO) is a class of methods for global optimization used to minimize expensive black-box functions. 
BO builds a probabilistic surrogate model of the objective and then determines future evaluations via an acquisition function. 
Applications of BO include robotic gait control, sensor set selection, and neural network hyperparameter tuning \cite{shahriari2016botutorial, snoek2012practical, frazier2018tutorial}.
BO is favored in these tasks because of its sample-efficient nature.
Achieving this sample-efficiency demands that BO balance exploration and exploitation. 
However, standard acquisition functions such as expected improvement (EI) are too greedy and perform little exploration. 
As a result, they perform poorly on multimodal problems \citep{hernandez2014predictive} and have provably sub-optimal performance in certain settings, e.g., bandit problems \cite{srinivas2009gaussian}. 
A key research goal in BO is developing less greedy acquisition functions \cite{shahriari2016botutorial}. 
Examples include predictive entropy search (PES) \cite{hernandez2014predictive} or knowledge gradient (KG) \cite{frazier2008knowledge}. 
\citet{lam2016bayesian} frame the exploration-exploitation trade-off as a balance between immediate and future rewards in a continuous state and action space Markov decision process (MDP). 
In this framework, \textit{non-myopic} acquisition functions are optimal MDP policies, and promise better performance by considering the impact of future evaluations up to a given BO budget (also referred to as the \textit{horizon}). 

\begin{figure*}[th!]  
    \centering
    \includegraphics[width=12cm, trim={0cm 0cm 0cm 0cm}]{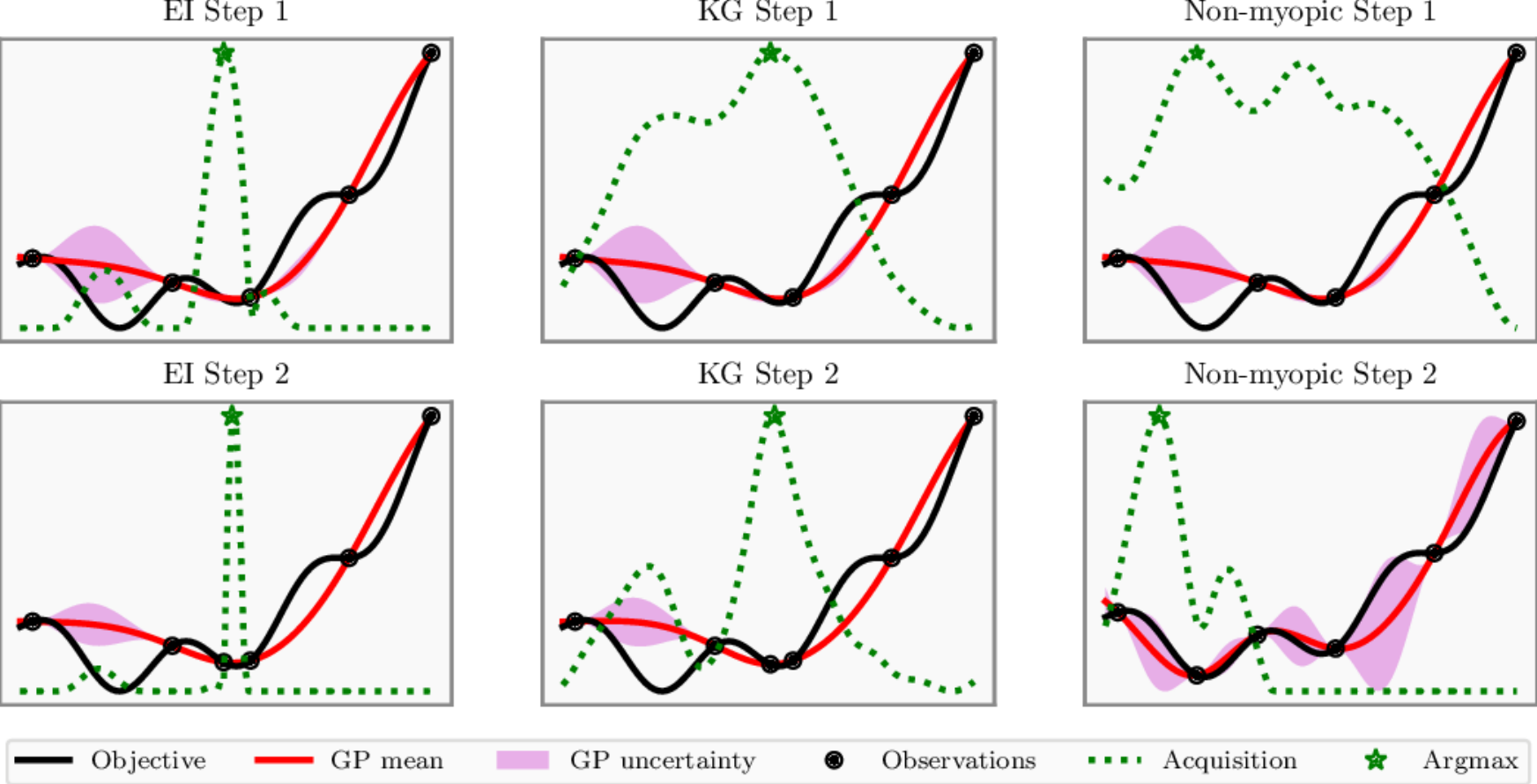}
    \caption{
        Comparing EI (\textit{left}), KG (\textit{middle}), and a non-myopic acquisition function (\textit{right}) on a carefully chosen objective. 
        We observe five values of $f(x) = \sin(20x) + 20(x - 0.3)^2$. 
        The GP has one promising minimum, but also a region of high uncertainty. 
        For each acquisition function, we perform two steps of BO. 
        EI will ignore the left region and instead greedily evaluate twice in a sub-optimal location. 
        KG is less greedy, but will nonetheless evaluate similarly to EI. 
        The non-myopic acquisition will properly evaluate the region of high uncertainty and identify the global minimum. 
    }
    \label{fig:eikglh}
\end{figure*}

Unfortunately, maximizing the MDP reward is an intractable problem as it involves solving an infinite-dimensional dynamic program \cite{powell2011adp}. 
Rollout is a popular class of approximate solutions in which future BO realizations and their corresponding values are simulated using the surrogate and then averaged.
This average defines a rollout acquisition function, which is considered the state-of-the-art in non-myopic BO. 
While more practical than the original MDP problem, rollout acquisition functions are still computationally expensive to the extent that suggesting the next evaluation can take several hours \cite{wu2019practical}.
This paper aims to make non-myopic BO more practical by reducing the time to suggest the next point from hours to seconds.
In particular, our main contributions are:
\begin{itemize}
    \item We compute rollout acquisition functions via Monte Carlo integration, and use variance reduction techniques ---quasi-Monte Carlo, common random numbers, and control variates--- to decrease the estimation error by several orders of magnitude.
    \item We introduce a coarser approximation to rollout acquisition functions through policy search. 
    In this setting, we use rollout to select the best acquisition function from a given set at each BO iteration. 
    This removes the need to optimize the rollout acquisition function, and makes rollout much more practical. 
    \item We provide experimental results for both rollout acquisitions and policy search. 
    The former suggests that rollout acquisition functions perform better on multimodal problems. 
    The latter shows that policy search performs as well as, if not better than, the best-performing one-step acquisition function. 
    \item We examine the impact of model mis-specification on performance and show that increasing the rollout horizon also increases sensitivity to model error.
\end{itemize}

\section{BACKGROUND}
\label{sec:background}
Many papers have been published in the sub-field of non-myopic BO over the last few years \cite{ginsbourger2009towards, gonzalez2016glasses, lam2017lookahead, lam2016bayesian, osborne2009gaussian, wu2019practical, yue2019lookahead}. 
Most of this recent research concerns rollout, in which future realizations of BO are simulated over horizon $h$ using the surrogate and averaged to determine the acquisition function. 
Rollout acquisition functions represent the state-of-the-art in BO and are integrals over $h$ dimensions, where the integrand itself is evaluated through inner optimizations, resulting in an expensive integral. 
The rollout acquisition function is then maximized to determine the next BO evaluation, further increasing the cost. 
This large computational overhead is evidenced by \citet{osborne2009gaussian}, who are only able to compute the rollout acquisition for horizon 2, dimension 1; and later by \citet{lam2016bayesian}, who use Gauss-Hermite quadrature in horizons up to five and see runtimes on the order of hours for small, synthetic functions \cite{wu2019practical}. 

Recent work focuses on making rollout more practical.
\citet{wu2019practical} consider horizon two and use a combination of Gauss-Hermite quadrature and Monte Carlo (MC) integration to quickly calculate the acquisition function and its gradient. 
Non-myopic active learning also uses rollout and recent work develops a fast implementation by truncating the horizon and selecting a batch of points to collect future rewards \cite{jiang2017efficient, jiang2018efficient}.

\begin{figure*}[t]
    \centering
    \includegraphics[width=15cm]{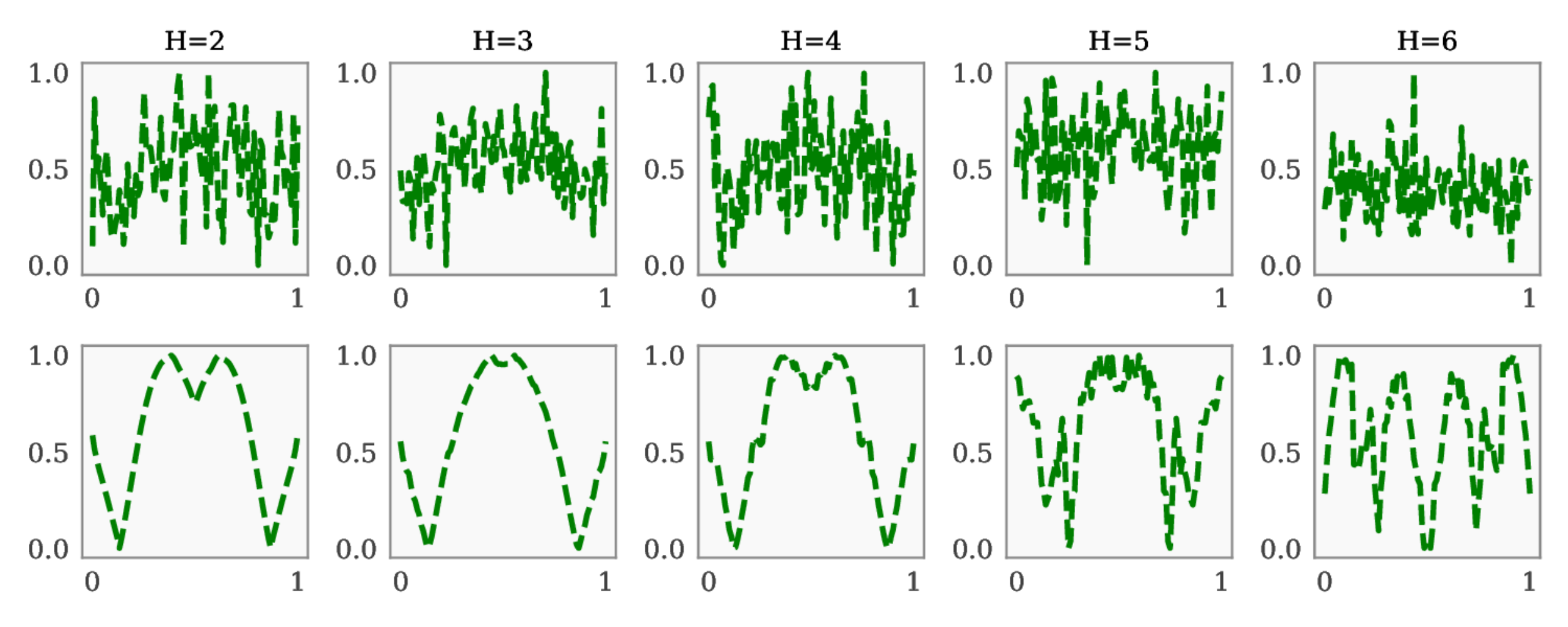}
    \caption{We calculate a standard non-myopic acquisition function in 1D with different values of $h$ using 50$h$ samples. (\textit{Top}) The results of a standard MC estimator. (\textit{Bottom}) The results of our estimator look far less noisy.}\label{fig:mc_comparison}
\end{figure*}

\textbf{Gaussian process regression and BO:} Suppose we seek a global minimum of a continuous objective $f(\bx)$ over a compact set $\Omega \subseteq \R^d$. 
If $f(\bx)$ is expensive to evaluate, finding a minimum should ideally be sample-efficient.
BO often uses a Gaussian process (GP) to model $f(\bx)$ from the data $\D_k = \{ (\bx_i,y_i) \}_{i=1}^k$. 
The next evaluation location $\bx_{k+1}$ is determined by maximizing an acquisition function $\Lambda(\bx \mid \D_k)$: $\bx_{k+1} =\argmax_{\Omega} \Lambda(\bx \mid \D_k)$. 

We place a GP prior on $f(\bx)$, denoted by $f \sim \GP(\mu,K)$, where $\mu:\Omega\to\R$ and $K:\Omega\times\Omega\to\R$ are the mean function and covariance kernel, respectively (see the supplement for examples). 
The kernel $K(\bx, \bx')$ correlates neighboring points, and may contain \textit{hyperparameters}, such as lengthscales that are learned to improve the quality of approximation \cite{rasmussen2006gaussian}. 
For a given $\D_k$, we define:
\[
    \by = \begin{pmatrix}y_1 \\\vdots\\ y_k\end{pmatrix},
    \,\,
    \bk(\bx) = \begin{pmatrix}K(\bx, \bx_1) \\\vdots\\ K(\bx, \bx_k)\end{pmatrix},
    \,\,
    \mK = \begin{pmatrix}\bk(\bx_1)^\top \\\vdots\\ \bk(\bx_k)^\top\end{pmatrix}.
\]
We assume $y_i$ is observed with Gaussian white noise: $y_i = f(\bx_i) + \epsilon_i$, where $\epsilon_i \sim \N(0, \sigma^2)$.  Given a GP prior and data $\D_k$, the resulting posterior distribution for function values at a location $\bx$ is the Normal distribution $\N(\mu^{(k)}(\bx ; \D_k), K^{(k)}(\bx,\bx; \D_k))$:
\begin{align*}
    \mu^{(k)}(\bx ; \D_k) =
        &\ \mu(\bx) + \bk(\bx)^\top(\mK+\sigma^2\mI_k)^{-1}(\by - \mu(\bx)), \\
    K^{(k)}(\bx,\bx; \D_k) &=
        K(\bx, \bx) - \bk(\bx)^\top(\mK+\sigma^2\mI_k)^{-1}\bk(\bx),
\end{align*}
where $\mI_k \text{ is the $k\times k$ identity matrix}$.

\textbf{Non-myopic Bayesian optimization:} Non-myopic BO frames the exploration-exploitation trade-off as a balance of immediate and future rewards. The strength of this approach is demonstrated in \figref{fig:eikglh}, in which we compare EI and KG to a non-myopic acquisition function on a carefully chosen objective. Its GP has a region of uncertainty on the left containing the global minimum and a local minimum at the origin. We run two steps of BO, updating the posterior each step. EI and KG behave greedily by sampling twice near the origin, while the non-myopic approach uses one evaluation to explore the uncertain region, subsequently identifying the global minimum and converging faster than either EI or KG.

\citet{lam2016bayesian} formulate non-myopic BO as a finite horizon dynamic program. 
We present the equivalent Markov decision process (MDP) formulation. We use standard notation from Puterman \cite{puterman2014mdp}: an MDP is the collection $<T, \bbS, \A, P, R>$. $T =  \{0, 1, \ldots, h-1\}$, $h < \infty$ is the set of decision epochs, assumed finite for our problem. The state space, $\bbS$, encapsulates all the information needed to model the system from time $t \in T$. 
$\A$ is the action space. Given a state $s \in \bbS$ and an action $a \in \A$, $P(s'| s, a)$ is the transition probability of the next state being $s'$. $R(s, a, s')$ is the reward received for choosing action $a$ from state $s$, and ending in state $s'$. 

A decision rule, $\pi_t : \bbS \rightarrow \A$, maps states to actions at time $t$. 
A policy $\pi$ is a series of decision rules $\pi = (\pi_0, \pi_1, \ldots, \pi_{h-1})$, one at each decision epoch. 
Given a policy $\pi$, a starting state $s_0$, and horizon $h$, we can define the expected total reward $V_h^\pi(s_0)$ as:
\[
    V_h^\pi(s_0) =  \E \bigg[\sum_{t=0}^{h-1} R(s_t, \pi_t(s_t), s_{t+1}) \bigg].
\]
In phrasing a sequence of decisions as an MDP, our goal is to find the optimal policy $\pi^*$ that maximizes the expected total reward, i.e., $\sup_{\pi \in \Pi} V_h^\pi (s_0)$, where $\Pi$ is the space of all admissible policies.

If we can sample from the transition probability $P$, we can estimate the expected total reward of any \textit{base policy} $\tilde \pi$ with MC integration:
\[
    V^{\tilde \pi}_h(s_0) \approx \frac{1}{N} \sum_{i=1}^N \bigg[\sum_{t=0}^{h-1} R(s^i_t, \tilde \pi_t(s^i_t), s^i_{t+1}) \bigg].
\]

Given a GP prior over data $\D_t$ with mean $\mu^{(t)}$ and kernel $K^{(t)}$, we model $h$ steps of BO as an MDP. 
This MDP's state space is all possible data sets reachable from starting state $\D_t$ with $h$ steps of BO. 
Its action space is $\Omega$; actions correspond to sampling a point in $\Omega$. Its transition probability and reward function are defined as follows. Given an action $\bx_{t+1}$, the transition probability from $\D_t$ to $\D_{t+1}$, where $\D_{t+1} = \D_t \cup \{(\bx_{t+1}, y_{t+1})\}$ is:
\begin{align*}
    &\ P(\D_t, \bx_{t+1}, \D_{t+1}) \sim \\
    &\ \N(\mu^{(t)}(\bx_{t+1} ; \D_t), K^{(t)}(\bx_{t+1}, \bx_{t+1}; \D_t)). 
\end{align*}
In other words, the transition probability from $\D_t$ to $\D_{t+1}$ is the probability of sampling $y_{t+1}$ from the posterior $\GP(\mu^{(t)}, K^{(t)})$ at $\bx_{t+1}$.
We define a reward according to EI \cite{jones1998ego}. 
Let $y_t^*$ be the minimum observed value in the observed set $\D_t$, i.e., $y_t^* = \min \{ y_0, \dots, y_t\}$. 
Then our reward is expressed as:
\begin{equation*}
    R(\D_t, \bx_{t+1}, \D_{t+1}) = ( y_t^* - y_{t+1})^+ \equiv  \max (y_t^* - y_{t+1},0).
\end{equation*}
EI can be defined as the optimal policy for horizon one, obtained by maximizing the immediate reward:
\begin{align*}
    \pi_{EI} &= \argmax_{\pi} V^{\pi}_1(\D_k) \\
    &= \argmax_{\bx \in \Omega} \E \Big[ (y_k^* - y_{\bx})^+ \Big] \equiv \argmax_{ \bx \in \Omega} \EI (\bx \mid \D_k),
\end{align*}
where the starting state is $\D_k$. 
In contrast, we define the non-myopic policy as the optimal solution to an $h$-horizon MDP. 
The expected total reward of this MDP can be expressed as:
\begin{align*}
\label{eq:hmdp}
    V_h^\pi(\D_k) &= \E \bigg[ \sum_{t=k}^{k + h-1} R(\D_t, \pi_t(\D_t), \D_{t+1}) \bigg] \\
    &= \E \bigg[ \sum_{t=k}^{k + h-1} (y^*_t - y_{t+1} )^+ \bigg] \nonumber.
\end{align*}
For $h>2$, the optimal policy is difficult to compute. 

\textbf{Rollout acquisition functions:} In the context of BO, rollout policies \cite{bertsekas1995dynamic}, which are sub-optimal but yield promising results, are a tractable alternative to optimal policies \cite{wu2019practical}. 
For a given current state $\D_k$, we denote our base policy $\tilde \pi = (\tilde \pi_0, \tilde \pi_1, \ldots ,\tilde \pi_{h-1})$. 
We introduce the notation $\D_{k,0}\equiv\D_k$ to define the initial state of our MDP and $\D_{k,t}$ for $1\leq t\leq h$ to denote the random variable that is the state at each decision epoch.  
Each individual decision rule $\tilde{\pi}_t$ consists of maximizing the base acquisition function $\Lambda$ given the current state $s_t=\D_{k,t}$,
\[
  \tilde \pi_t = \argmax_{\bx \in \Omega} \Lambda(\bx \mid \D_{k,t}). 
\]
Using this policy, we define the non-myopic acquisition function $\Lambda_h(\bx)$ as the rollout of $\tilde \pi$ to horizon $h$ i.e., the expected reward of $\tilde{\pi}$ starting with the action $\tilde{\pi}_0=\bx$:
\[
    \Lambda_{h}(\bx_{k+1}) := \E \bigg[ V^{\tilde \pi}_h (\D_k \cup \{(\bx_{k+1}, y_{k+1})\}) \bigg] ,
\]
where $y_{k+1}$ is the noisy observed value of $f$ at $\bx_{k+1}$. $\Lambda_h$ is better than $\Lambda$ in expectation for a correctly specified GP prior and for any acquisition function. This follows from standard results in the MDP literature:

\begin{definition}
\citet{bertsekas1995dynamic}: A policy $\pi$ is sequentially consistent if, for every trajectory generated from any $s_0$:
\[
(s_0, a_0), (s_1, a_1), \dots, (s_{h-1}, a_{h-1}),
\]
$\pi$ generates the following trajectory starting at $s_1$:
\[
(s_1, a_1), (s_2, a_2) \dots, (s_{h-1}, a_{h-1}).
\]
\end{definition}

Sequential consistency requires the trajectory generated from applying $\pi$ at $s_i$ for horizon $h-i$ to be a sub-trajectory of the trajectory generated from applying $\pi$ at $s_0$ for horizon $h$. Acquisition functions are sequentially consistent so long as we consistently break ties if they have multiple maxima ---though this will not occur generically. Sequential consistency guarantees that rollout does at least as well as its base policy in expectation:

\begin{theorem}\label{theorem:mdp_rollout}
\citet{bertsekas1995dynamic}:
A rollout policy $\pi_{roll}$ does as least as well as its base policy $\tilde \pi$ in expectation if $\tilde \pi$ is sequentially consistent i.e.,
\[
V^{\pi_{roll}}_h(s_0) \geq V^{\tilde \pi}_h(s_0).
\]
\end{theorem}

Thus, rolling out any acquisition function will do at least as well in expectation as the acquisition function itself.

Unfortunately, while rollout is tractable and conceptually straightforward, it is computationally demanding. 
To rollout $\tilde \pi$ once, we must run $h$ steps of BO with $\Lambda$. 
Many such rollouts must then be averaged to reasonably estimate $\Lambda_h$, which is an $h$-dimensional integral. 
Estimation can be done either through explicit quadrature or MC integration, and is the primary computational bottleneck of rollout. 
Our paper significantly lowers the computational burden of rollout through MC variance reduction and a fast policy search method that avoids optimizing $\Lambda_h$. 
We detail these methods in the next section.

\section{METHODS}
\label{sec:methods}
In the context of rollout, MC estimates the expected reduction over $h$ steps of BO using base policy $\tilde \pi$:
\begin{align*}
 \Lambda_{h}(\bx_{k+1}) = &\ \E \bigg[ V^{\tilde \pi}_h (\D_k \cup \{(\bx_{k+1}, y_{k+1})\}) \bigg] \\
\approx &\ \frac{1}{N} \sum_{i=1}^N \sum_{t=k}^{k + h-1} (y^*_t - y_{t+1} )^+.
\end{align*}
The distribution for $y_{t+1}$ is a normal distribution whose mean and variance are determined by rolling out $\tilde \pi$ to horizon $t$ and examining the posterior GP:
\begin{equation}
    \label{eq:implicit_distribution}
    \begin{aligned}
        y_{t+1} \sim &\ \N(\mu^{(t)}(\bx_{t+1} ; \D_t), K^{(t)}(\bx_{t+1}, \bx_{t+1}; \D_t)), \\
        \bx_{t+1} = &\ \pi ( \D_{t}) = \argmax_{\bx \in \Omega} \Lambda(\bx \mid \D_{t}).
    \end{aligned}
\end{equation}
A sample path in this context may be represented as the sequence $(\bx_k, y_k), (\bx_{k+1}, y_{k+1}), \dots, (\bx_{k+h}, y_{k+h})$ produced by Equation~\ref{eq:implicit_distribution}. We parameterize the vector of $y$ values, $\by$, with an $h$-dimensional vector $\mathbf{z}$ drawn from $\N(0, \mI_h)$. 
$y_{t+1}$ is distributed according to $\N(\mu^{(t)}(\bx_{t+1} ; \D_t), K^{(t)}(\bx_{t+1}, \bx_{t+1}; \D_t))$, so we map $\mathbf{z}_{t+1}$ to $y_{t+1}$ by a simple scale-and-shift. 
This map is done sequentially from time step one to time step $h$. 
MC integration thus involves sampling $N$ times from $\N(0, \mI_h)$, mapping each of the samples, and averaging. 
The mapping step is equivalent to applying our rollout policy $\tilde \pi$, and is the dominant cost of integration. 

Compared to other quadrature schemes, MC is well-suited to high-dimensional integration. 
MC converges at a rate of $\sigma / \sqrt{N}$, the standard deviation of the MC estimator, where $\sigma$ is the sample variance and $N$ is the total number of samples. 
MC's primary drawback is slow convergence. 
Increasing precision by an order of magnitude requires two orders of magnitude more samples. If $\sigma$ is high, many samples may be required to converge. 
In this section, we focus on two strategies that significantly decrease the overhead of rollout: variance reduction and policy search.

\subsection{Variance reduction}
\begin{figure}[t!]
    \centering
    \includegraphics[width=0.48\textwidth, trim={0cm 1cm 0cm 0cm}]{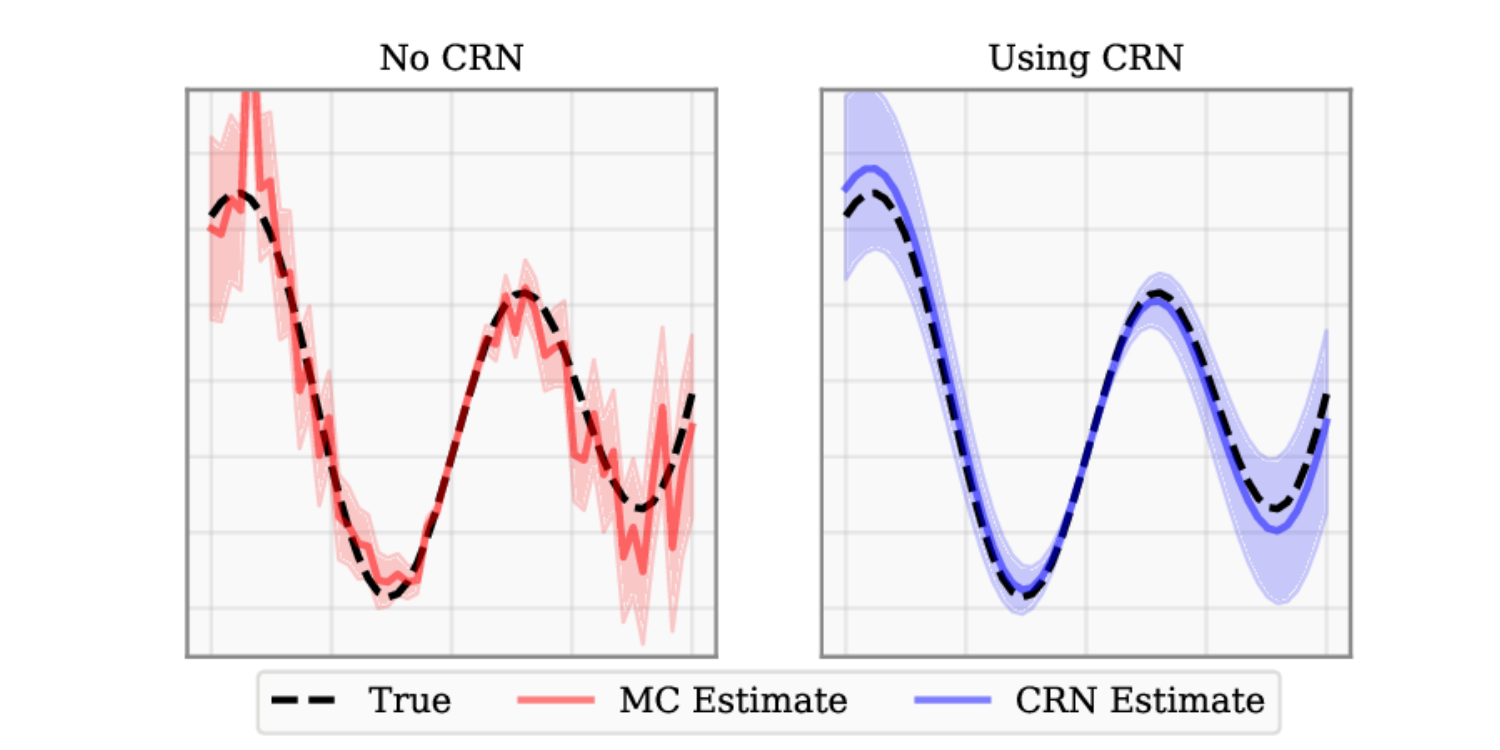}
    \caption{We estimate a function via MC. (\textit{Left}) Standard MC is noisy, and the estimate's argmin is not the function's. (\textit{Right}) Using CRN makes a significant difference. We shade point-wise variances above and below. Note that they are unchanged; CRN only affects the estimate's smoothness, which in this case yields the correct argmin.}
    \label{fig:crn}
\end{figure}

Variance reduction is a class of methods that improve convergence by decreasing the variance of the estimator. 
Effective variance reduction methods can reduce $\sigma$ by orders of magnitude. 
We use a combination of quasi-Monte Carlo, common random numbers, and control variates, which significantly reduces the number of MC samples needed, as evidenced by Figure~\ref{fig:mc_comparison}. 

\textbf{Quasi-Monte Carlo (QMC):} Instead of sampling directly from the probability distribution, QMC instead uses a low-discrepancy sequence as its sample set. 

\begin{theorem} \citet{caflisch1998monte}:
QMC converges at a rate bounded above by $\log(N)^h / N$, where $N$ is the number of samples and $h$ is the integral's dimension. 
\end{theorem}\label{theorem:qmc_convergence}

This bound stems from the well-known Koksma-Hlawka inequality \cite{caflisch1998monte}, and is roughly linear for large $N$. 
In practice, this bound is often loose and convergence proceeds faster \cite{papageorgiou2003sufficient}. 
QMC is inapplicable if a low-discrepancy sequence does not exist for the target distribution. 

In our case, the distributions we integrate over are Normal, for which low-discrepancy sequences do exist. We generate low-discrepancy Sobol sequences in the $h$-dimensional uniform distribution $\mathcal{U}[0,1]^h$ and map them to the unit multivariate Gaussian via the Box-Muller transform. 
This yields a low-discrepancy sequence for $\N(0, \mI_h)$. 
Recall the parameterization of samples from $\N(0, \mI_h)$ to sample rollout trajectories in Equation~\ref{eq:implicit_distribution}. 
We apply QMC by replacing the unit multivariate Gaussian samples with our low-discrepancy sequence. 

\begin{figure*}[t]
    \centering
    \includegraphics[width=\textwidth, trim={0cm 2.5cm 0cm 0cm}]{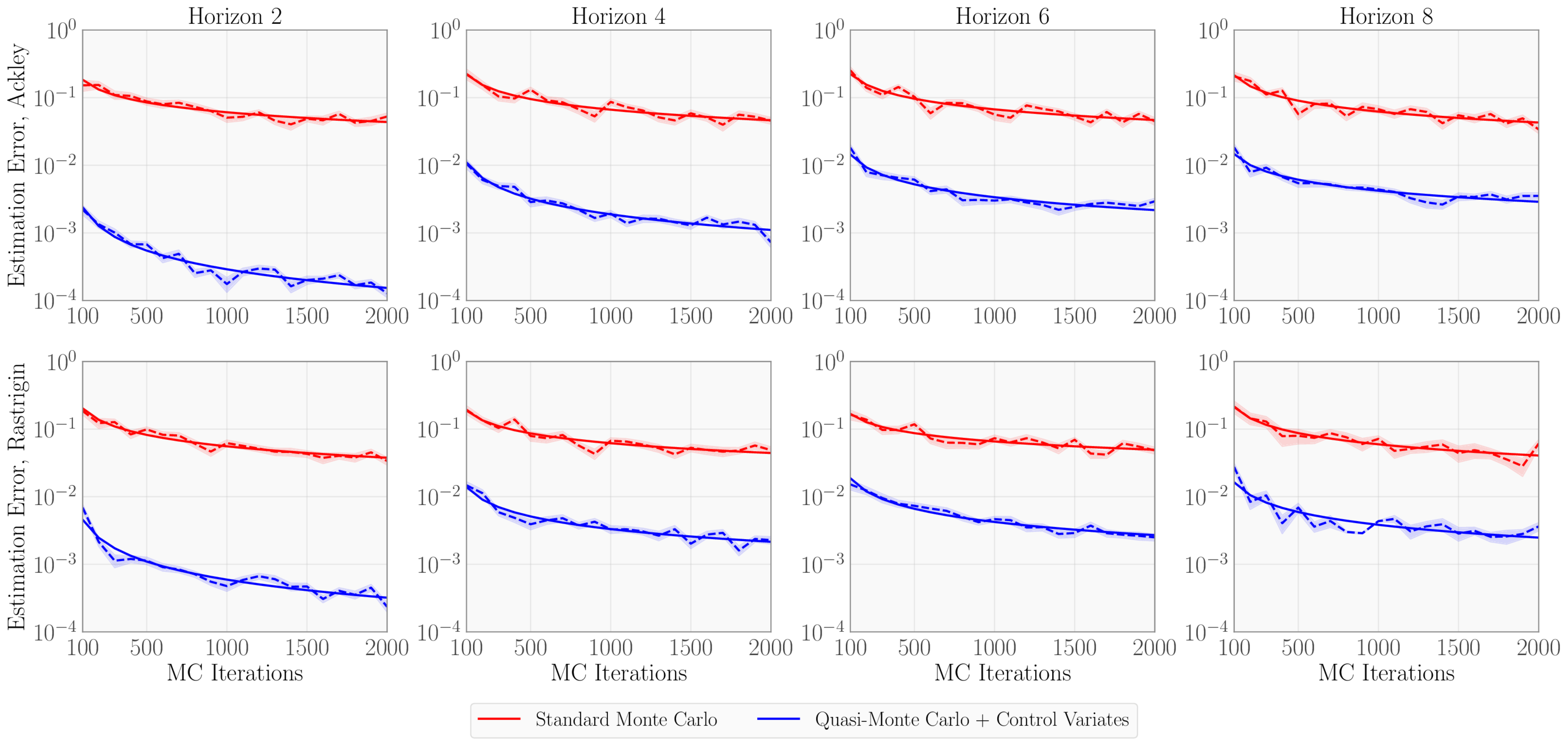}
    \caption{The estimation errors of MC (\textit{red}) and our reduced-variance estimator (\textit{blue}).}\label{fig:var_red}
\end{figure*}

\textbf{Common random numbers (CRN):} CRN is used when estimating a quantity to be optimized over parameter $\bx$, and is implemented by using the same random number stream for all values of $\bx$. 
CRN does not decrease the point-wise variance of an estimate, but rather decreases the covariance between two neighboring estimates, which smooths out the function. 
Consider estimating $\E_y [f(y, \bx_1)]$ and $\E_y [f(y, \bx_2)]$ for two points $\bx_1$ and $\bx_2$ using $N$ samples, with variances $\sigma_1 / N$ and $\sigma_2 / N$ respectively. 
We define the differences $\delta$ and $\hat \delta$ as: 
\begin{align*}
    \delta =&\ \E_y [f(y, \bx_1)] - \E_y [f(y, \bx_2)], \\
    \hat \delta =&\ \E_y [f(y, \bx_1) - f(y, \bx_2)], \\
    \Var[\delta] =&\ (\sigma_1^2 + \sigma_2^2) / N, \\
    \Var[\hat \delta] =&\ (\sigma_1^2 + \sigma_2^2 - 2\,\Cov[f(y, \bx_1), f(y, \bx_2)]) / N.
\end{align*}
$\hat \delta$ uses the same number stream for $\bx_1$ and $\bx_2$. 
If $\bx_1$ and $\bx_2$ are close and $f$ is continuous in $\bx$, $\Cov[f(y, \bx_1), f(y, \bx_2)]) > 0$, implying $\Var[\hat \delta] < \Var[\delta]$. 
This increased consistency between neighboring points improves optimization accuracy, as seen in Figure~\ref{fig:crn}. 

\textbf{Control variates:} The general idea behind control variates is to find a covariate $g(y)$ with the same distribution as $f(y)$ and a known mean. The quantity $c(y) = f(y) - \beta g(y)$, known as a regression control variate (RCV), is estimated, and then de-biased afterwards. 

\begin{theorem}
Consider the estimator $\E[c(y)] = \E[f(y) - \beta g(y)]$. Let $\Var[f(y)], \Var[g(y)] = \sigma_f, \sigma_g$. $g(y)$ is sufficiently correlated with $f(y)$ if:
\[
 \beta^2\sigma_g^2 - 2\beta \Cov[f(y), g(y)] < 0.
\]

If $g(y)$ is sufficiently correlated with $f(y)$, then the estimator $\E[c(y)]$ is strictly more accurate than $\E[f(y)]$ i.e., $\Var[c(y)] < \Var[f(y)]$.
\end{theorem}\label{theorem:controlvariate_convergence}

The optimal value that minimizes the variance of $c(y)$ is $\beta = \Cov[f(y), g(y)] / \sigma_h^2$. 
In practice, both $ \Cov[f(y), g(y)]$ and $\sigma_h^2$ must be estimated. 
This formula generalizes to multiple control variates (included in the supplementary). 
We note that variates whose derivatives are also correlated with the derivatives of $f(y)$ yield superior performance, especially when using QMC \cite{hickernell2005control}.   

In the context of BO, we opt to use covariates derived from existing acquisition functions with known means. 
EI and PI are straightforward options. 
We expect their value to be at least somewhat correlated with the value of the rollout acquisition function; a promising candidate point should ideally score highly among all acquisition functions, and vice-versa. 
We will demonstrate the effectiveness of these variates in Section~\ref{sec:experiments}. 
EI and PI are defined as the expectation of a random variable of the form 
$\Lambda(\bx) =  \mathbb{E}[\gamma(y | \bx)] = \int_{-\infty}^{\infty} \gamma(y|\bx) p(y|\bx) dy$:
\begin{itemize}
    \item \textit{Probability of improvement (PI)}  
\[
    \gamma(y | \bx) =\left\{
                \begin{array}{ll}
                  1 \;,\; y < y^*\\
                  0 \;,\; y \geq y^*.
                \end{array} \right\} 
\]
    \item \textit{Expected improvement (EI)} 
\[
    \gamma(y | \bx) =\left\{
                \begin{aligned}
                \begin{array}{ll}
                  y^* - y \;,\; & y < y^*\\
                  0 \;,\; & y \geq y^*.
                \end{array} 
                \end{aligned} \right\}
\]
\end{itemize}

\subsection{Policy search}
While we have dramatically lowered the cost of evaluating the rollout acquisition function, there still remains the problem of its optimization. 
\citet{wu2019practical} use the reparameterization trick to estimate the gradient of EI for horizon two and use stochastic gradient descent to maximize it. 
However, their method does not immediately extend to horizons larger than two.

Policy search is an alternative method for approximately solving MDPs, in which a best performing policy is chosen out of a (possibly infinite) set of policies $\Pi = \{ \pi_1, \pi_2, \dots \}$ \cite{bertsekas1995dynamic}. 
It is performed either by computing the expected reward for each policy in the set, or using a gradient-based method to maximize the expected reward given a parameterization of the policy set. 
In this paper, we use a finite policy set:
\[
    \Pi_{ps} = \{ \pi(\D_k) \ | \ \pi(\D_k) := \argmax \Lambda(\bx | \D_k) \in A \},
\]
where $A$ is any arbitrary set of acquisition functions. We then select the best-performing policy:
\[
    \pi_{ps} = \argmax_{\pi_i \in \Pi_{ps}} V_h^{\pi_i}(\D_k).
\]
A less formal explanation follows: at every step of BO, we roll out each acquisition function in $A$ on its argmax, and use the one with the highest $h$-step reward. 
A related approach by \citet{hoffman2011portfolio} employs a bandit strategy to switch between different acquisition functions. 
Our policy search method does not maximize the rollout acquisition function, and is thus significantly faster, though it likely reduces performance. 
Experiments in Section~\ref{sec:experiments} suggest that our policy search method performs at least as well as the best-performing acquisition in $\Pi_{ps}$. 

\section{EXPERIMENTS AND DISCUSSION}
\label{sec:experiments}
\begin{figure*}[!ht]
    \centering
    \includegraphics[width=\textwidth, trim={0cm 1cm 0cm 0cm}]{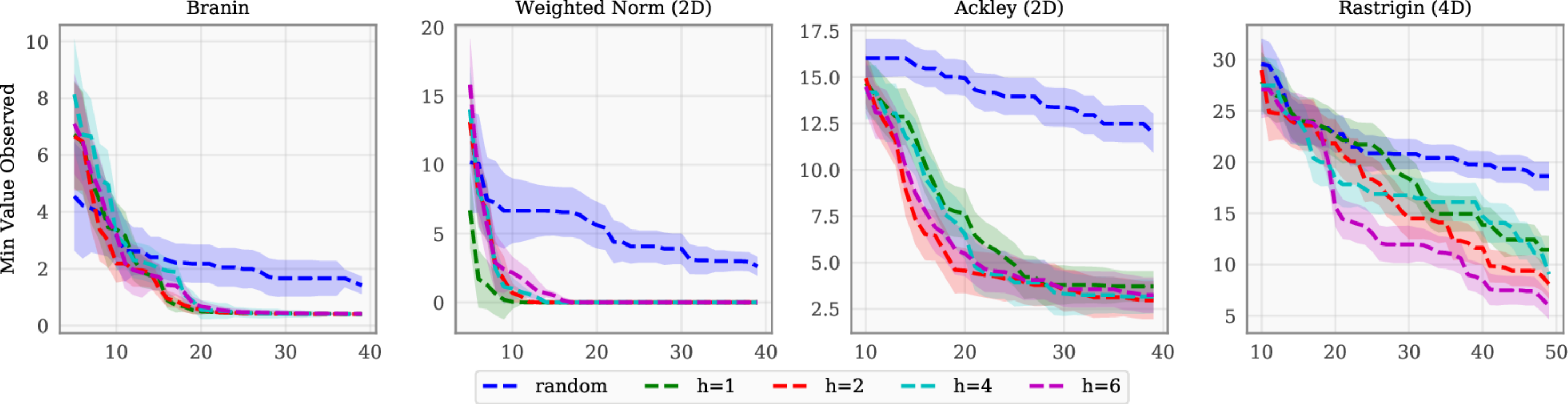}
    \caption{Empirically, looking longer horizons only seems to help on multimodal functions. On unimodal functions (not necessarily convex), there is little to no performance gain.}\label{fig:synthetic}
\end{figure*}

Unless otherwise stated, we use a GP with the Mat\'ern $5/2$ ARD kernel \cite{snoek2012practical} and learn its hyperparameters via maximum likelihood estimation \cite{rasmussen2006gaussian}. 
When rolling out acquisition functions, we maximize them with L-BFGS-B using five restarts, selected by evaluating the acquisition on a Latin hypercube of $10d$ points and picking the five best. 
EI is used as the base rollout policy except for in the policy search experiments. 
All synthetic functions are found in the supplementary. 
Code to reproduce our experiments is found at \url{https://github.com/ericlee0803/lookahead_release}. 

\textbf{Variance reduction experiments:} We compare the estimation error and convergence rate between the standard MC estimator and our estimator. 
We take $2d$ random points in the domain and evaluate the Ackley and Rastrigin functions in $2$D and $4$D, respectively. 
We roll out EI for horizons $2$, $4$, $6$, and $8$, and calculate the variance of both estimators for MC sample sizes in $[100, 200, 300, \dots, 2000]$, using $50$ trials each. 
We take the ground truth to be estimation with $10^{4}$ samples. 
The mean error of the standard MC estimator (\textit{red}) and our reduced-variance estimator (\textit{blue}) are plotted with dotted lines in Figure \ref{fig:var_red}, with standard error shaded above and below. We also compute a best-fit line to each mean error, which is plotted with a solid line. 

\begin{table}[!ht]
    \centering
    \small
    \setlength\tabcolsep{4pt}
    \caption{Estimated convergence rate and error reduction $\sigma / \hat \sigma$ for the standard MC estimator and our estimator}
    
     \begin{tabular}{|c c | c c c |} 
     \hline
     Objective & Horizon & MC Rate &  Our Rate & $\sigma / \hat \sigma$ \\ [0.5ex] 
     \hline \hline
    Ackley    & $2$ & $0.53$ & $0.95$ & $410$ \\ 
    Ackley    & $4$ & $0.52$ & $0.82$ & $63$  \\
    Ackley    & $6$ & $0.55$ & $0.64$ & $28$  \\
    Ackley    & $8$ & $0.53$ & $0.54$ & $26$  \\
    Rastrigin & $2$ & $0.56$ & $0.90$ & $150$ \\ 
    Rastrigin & $4$ & $0.48$ & $0.63$ & $31$  \\
    Rastrigin & $6$ & $0.47$ & $0.68$ & $30$  \\
    Rastrigin & $8$ & $0.42$ & $0.64$ & $25$  \\
    \hline
    \end{tabular}
    \label{tab:var_red_table}
\end{table}

Table \ref{tab:var_red_table} summarizes our experimental results, and includes our estimates for the convergence rate of both estimators and the relative reduction in estimation error $\sigma / \hat \sigma$. Our estimator has significantly lower estimation error ---the maximum reduction in estimation error we achieve is a factor of $410$. 
Standard MC clearly converges at a $N^{-1/2}$ rate. Our estimator converges like $N^{-1}$ for smaller horizons, but its convergence rate drops as $h$ increases. 
This is due to QMC's $\log(N)^h / N$ convergence. $N$ is not large enough for longer horizons to exhibit $N^{-1}$ convergence; increasing it past $2000$ should yield $N^{-1}$ convergence. 

Another trend is the increase in estimation error as the horizon increases, which is expected given that the dimensionality of the underlying integral increases. 
Fortunately, the error seems to increase only linearly ---and by a small constant--- rather than exponentially, suggesting that MC samples proportional to $h$ is sufficient to achieve a high quality of approximation. 
Finally, the reduction in estimation error levels off to around a factor of $25$, suggesting that the correlation between the rollout acquisition function and our control variates decreases when $h$ increases. 
We include an ablation study in the supplement to quantify the individual contributions of QMC and control variates. 

A factor of 25 error reduction is still significant; the standard MC estimator would need $625$ times more samples to achieve comparable accuracy. 

\begin{figure*}[!ht]
    \centering
    \includegraphics[width=\textwidth, trim={0cm 0cm 0cm 0cm}]{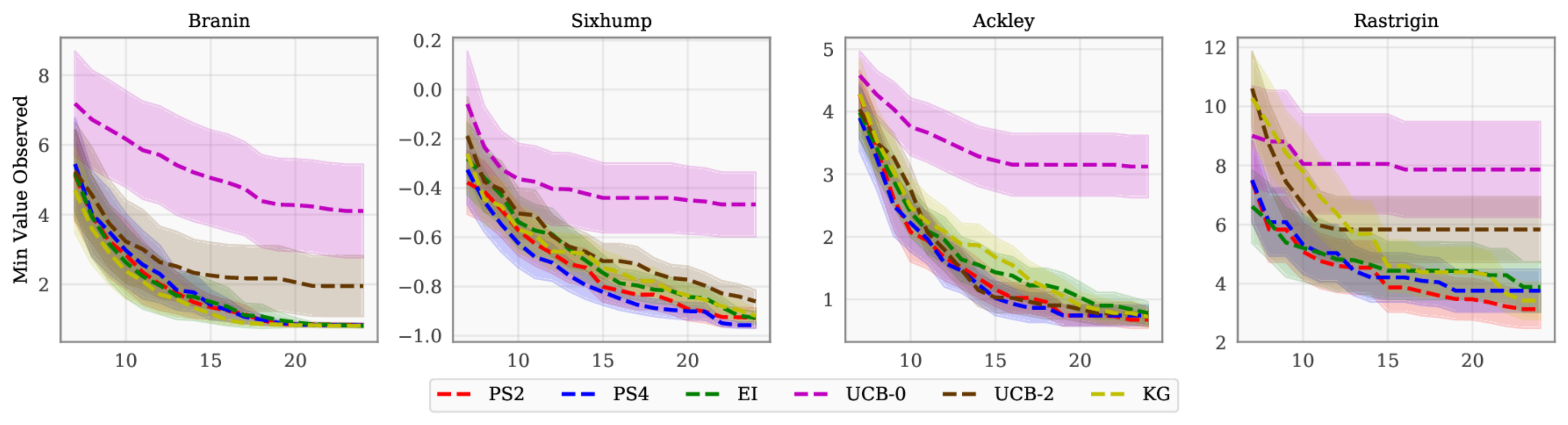}
    \includegraphics[width=\textwidth, trim={0cm 1cm 0cm 0cm}]{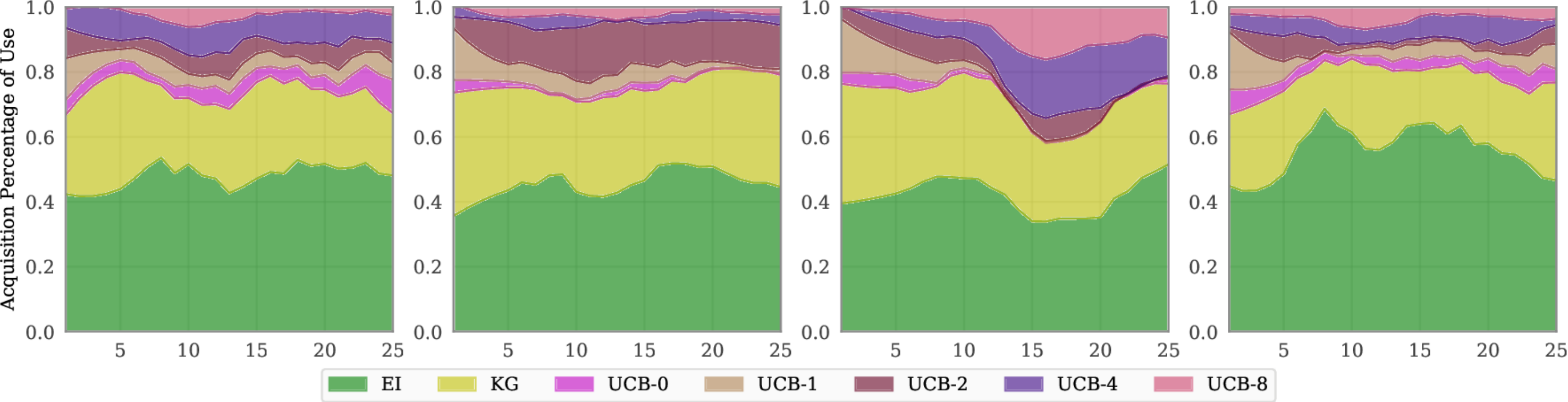}
    \caption{ (\textit{Top}) Policy search performs at least as well as the best acquisition function, if not better. (\textit{Bottom}) For each of the corresponding objectives, we plot the percentage of use of each acquisition function per iteration for PS$4$. EI and KG are chosen more often than any of the UCB acquisition functions. The worst-performing acquisition, UCB-$0$, is chosen the least, suggesting correlation between an acquisition's performance and its percentage of use. }\label{fig:policysearch}
\end{figure*}

\textbf{Full rollout on synthetic functions:} We roll out EI for $h=2, 4$, and $6$ on the Branin, weighted-two-norm ($2$D), Ackley ($2$D), and Rastrigin ($4$D) functions in Figure~\ref{fig:synthetic} using $200h$ MC samples, and compare to both standard EI and random search. 
To optimize the acquisition functions quickly, we employ the following strategy: we evaluate the acquisition function on a Sobel sequence of size $10d$, as well as an additional point which is the argmax of EI. 
We then use this as an initial design and run BO for $50d$ more iterations. We run $50$ iterations for each horizon and provide random search as a baseline. 
The mean results and the standard error are plotted in Figure~\ref{fig:synthetic}. 

On the Branin function, all horizons performed comparably and converge in $20$ iterations. 
Rollout performed best on the Ackley and Rastrigin functions, which are multimodal. 
On the weighted norm function, which is strongly convex, EI converges within $10$ iterations, and looking ahead further yielded poorer results. 
These results suggest that more exploratory acquisitions are needed for a multimodal objective, whereas more exploitative acquisitions suffice for reasonably simple objective functions. 

\textbf{Policy search:} We consider policy search (PS) with an acquisition set of EI, KG, and Upper Confidence Bound (UCB--$\kappa$) for $\kappa \in \{0, 1, 2, 4, 8\}$ \cite{snoek2012practical}, which contains acquisitions that tend towards both exploitation and exploration. 
We run policy search for horizons $2$ and $4$ on the Branin, Sixhump, Ackley, and Rastrigin synthetic functions, all in $2$D, using $200h$ MC samples. 

All acquisition functions are maximized via L-BFGS-B with five random restarts, except for KG, which uses grid search of size $900$. 
The mean results and standard error over $50$ trials are plotted in Figure~\ref{fig:policysearch}, in which policy search for horizons $2$ and $4$, labeled PS$2$ and PS$4$ respectively, do better or on par with the best-performing acquisition function. 
This robustness is a key strength of policy search, as the performance of each acquisition function
is often problem-dependent.
\begin{figure*}[ht]
    \centering
    \includegraphics[width=14cm, trim={0cm 0.85cm 0cm 0cm}]{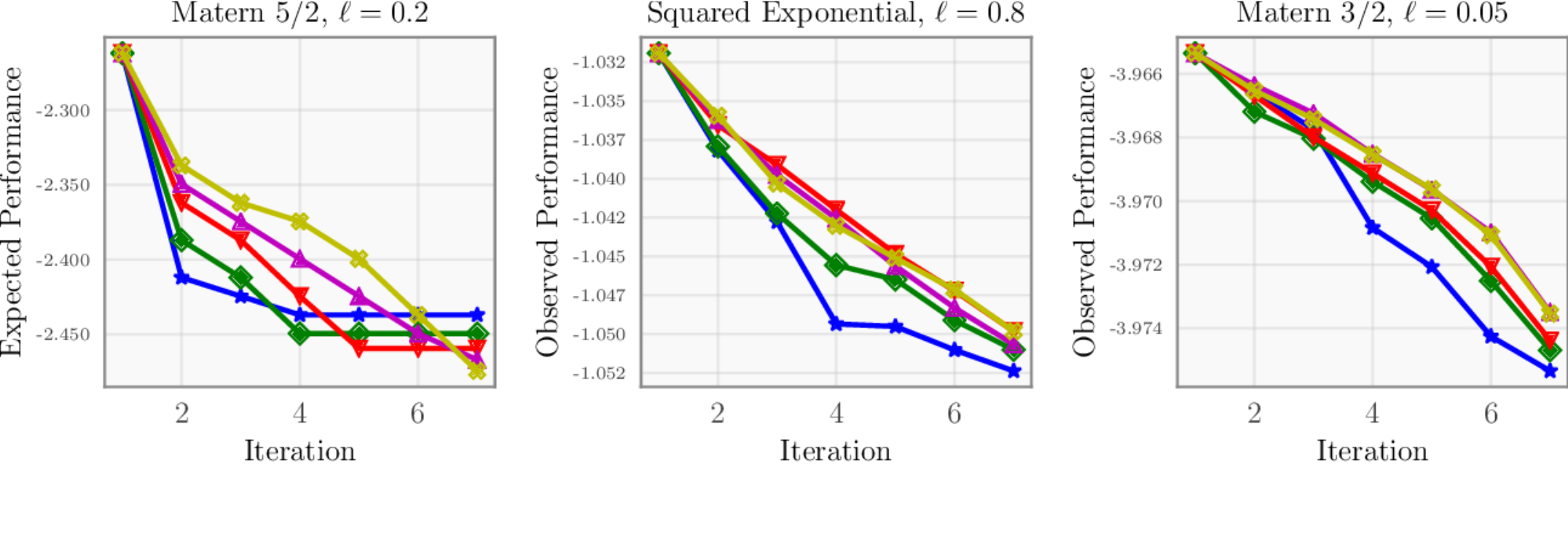}
    \includegraphics[width=8cm, trim={0cm 0cm 0cm 0cm}]{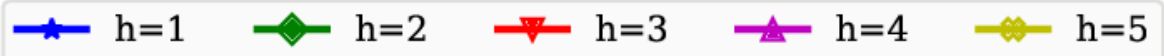}
    \caption{
    (\textit{Left}) The expected performance of EI-based rollout for $h$ = $1$, $2$, $3$, $4$, and $5$. 
    (\textit{Middle, Right}) The observed performance of rollout, given model error in the form of a smoother and less smooth kernel, respectively. 
    When the model has large error, the resulting performance of non-myopic policies can be reversed from the expected performance.
    }
    \label{fig:ModelMismatch}
\end{figure*}
We also examine the choice of acquisition function as a function of iteration. 
The percentage of use of each acquisition function is shown for its corresponding objective, and for plotting purposes we smooth the percentage with a box filter of size five. 
EI and KG are chosen more often than the other acquisition functions. 
UCB-$0$, the worst performing method representing a pure exploitation policy, is chosen significantly less than others, while UCB-$2$ was chosen the most often out of the UCB family. 
Of particular interest is the Ackley function (third column, Figure \ref{fig:policysearch}): when UCB-$2$ starts to outperform the other acquisitions functions, a clear spike in its percentage of use is seen in the corresponding histogram. 

\textbf{Neural architecture search (NAS) benchmark:}
We run policy search on the NAS tabular benchmarks in \cite{klein2019tabular}, which are an exhaustive set of evaluated configurations for multi-layer perceptrons trained on different datasets. We optimizer the perceptrons' layer sizes, batch size, and training epochs with 60 iterations of BO for each dataset and plot average regret in Figure \ref{fig:nas}. For space's sake, we describe the search space and the regret metric in the supplement. Policy search for $h=2$ and $h=4$ performs better than EI, UCB0, UCB2, and KG. 
\begin{figure}[]
    \centering
    \includegraphics[width=0.35\textwidth, trim={0cm 0cm 0cm 0cm}]{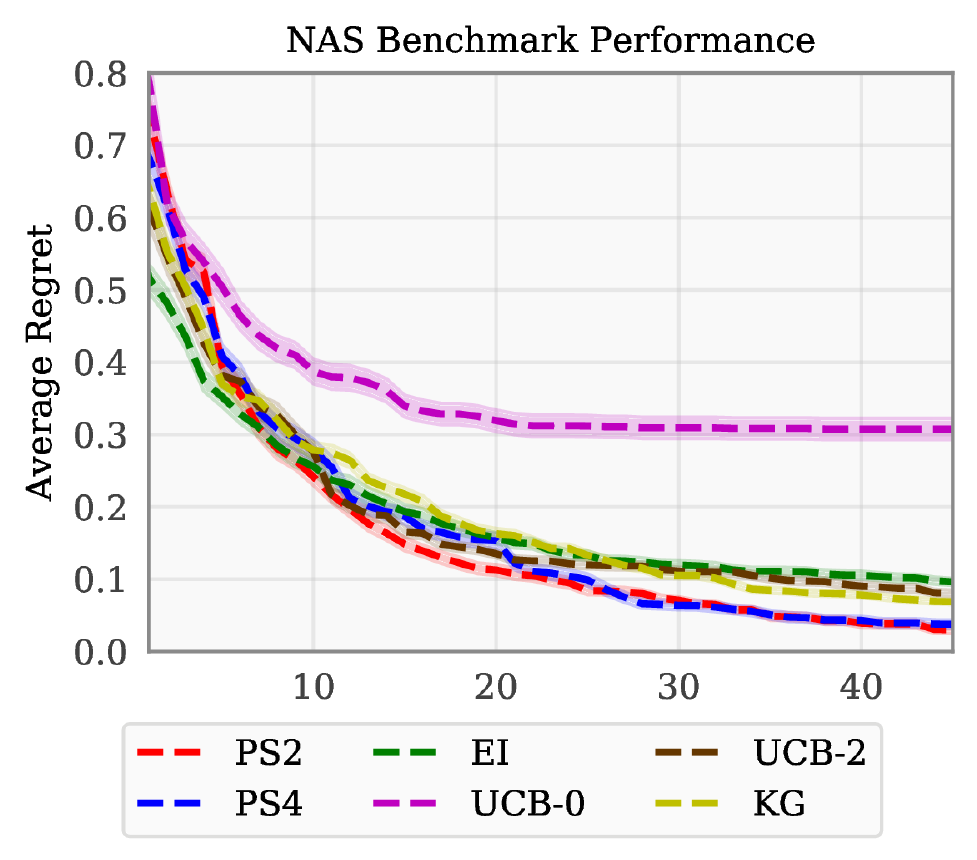}
    \caption{We compare the performance of policy search for horizons 2 and 4 in red and blue, respectively, with that of EI, UCB0, UCB2, and KG. PS2 and PS4 outperform the others after about 20 iterations. }
    \label{fig:nas}
\end{figure}

\textbf{The impact of model mis-specification:} We believe that any probabilistic model only supports a limited horizon due to the effects of model error.
Errors in the GP model result in errors to the MDP transition probabilities, which grow as they are propagated through time.
This likely renders long-horizon rollout ineffectual. 
We support this hypothesis by comparing the performance of policies in the MDP setting they were designed in with their observed performance on objectives drawn from a different MDP. 

We do this by drawing objective functions from a GP with fixed kernel every step of BO.
More concretely, evaluating the objective function at any point $\bx$ is performed by sampling from the GP posterior distribution at $\bx$. 
Because policies are designed to maximize this MDP's reward for a fixed horizon and because the objective is drawn from the MDP itself, policies looking further ahead perform better 
by definition.
We then draw objectives from a GP using a \textit{different} kernel with a \textit{different} lengthscale, and check if policies looking further ahead still perform better. 

We rollout EI in $1$D with a budget of seven and we model our objective with a GP using the Mat\'ern $5/2$ kernel with $\ell=0.2$.
In \figref{fig:ModelMismatch}, the left panel depicts expected performance of rollout for $h$ = $1$, $2$, $3$, $4$, and $5$. 
The middle and right panels depict observed performance of rollout when the we sample objectives from a GP that has a far smoother kernel (Square Exponential with $\ell=0.8$) and far less smooth kernel (Mat\'ern $3/2$ with $\ell=0.05$), respectively. 
All plots use $2000$ replications to achieve high accuracy. 

The result is perhaps unsurprising; the ranking of the observed performance of policies is reversed with that of the expected performance. 
Myopic BO performed the best; more generally, policies with shorter horizons outperformed those with longer horizons. 
This demonstrated sensitivity to model error suggests non-myopic BO must carefully strike a balance between model accuracy and horizon, and justifies use of modest horizons over the full BO budget. 
This confirms experimental results by \citet{yue2019lookahead}, who suggest looking ahead to short horizons is preferable to long horizons in practice.  

\section{CONCLUSION}
\label{sec:conclusion}
We have shown that a combination of quasi-Monte Carlo, control variates, and common random numbers significantly lowers the overhead of rollout in BO. 
We have introduced a policy search which further decreases computational cost by removing the need to maximize the rollout acquisition function. 
Finally, we have illustrated the penalties incurred by using inaccurate GP models in the non-myopic setting. 

This work raises several interesting research directions. Decreasing the variance of our estimator may be possible with additional variance reduction methods such as stratified or antithetic sampling. 
Developing a more comprehensive policy search space, such as a parameterized set of all convex combinations of acquisition functions, may further strengthen the policy search performance.  

\bibliography{references.bib}
\bibliographystyle{abbrvnat}

\appendix
\section*{Appendix}
\label{sec:appendix}

\section{KERNELS}
The kernel functions we use in this paper are the squared exponential (SE) kernel, Mat{\'e}rn 5/2 kernel, and Mat{\'e}rn 3/2 kernel, respectively:
\begin{align*}
    K_{\text{SE}}(r)  &= \alpha^2 \exp\left(-\frac{r^2}{2\ell^2} \right), \\
    K_{\text{5/2}}(r) &= \alpha^2\left( 1 + \frac{\sqrt{5}}{\ell} + \frac{5}{3\ell^2}\right) \exp\left(-\frac{\sqrt{5}r}{\ell} \right), \\
    K_{\text{3/2}}(r) &= \alpha^2\left( 1 + \frac{\sqrt{3}}{\ell} \right) \exp\left(-\frac{\sqrt{3}r}{\ell} \right),
\end{align*}
where $r = \| \bx - \bx' \|_2$.

\section{ACQUISITION FUNCTIONS}
PI, EI, and UCB-$\kappa$ have the closed forms:
\begin{align*}
    &\Lambda_{PI}(\bx) = \Phi(\frac{y(\bx) - y^*}{\sigma(\bx)}). \\
    &\Lambda_{EI}(\bx) = (y(\bx) - y^*) \Phi(\frac{y(\bx) - y^*}{\sigma(\bx)}) \\
    & \quad + \sigma(\bx) \phi(\frac{y(\bx) - y^*}{\sigma(\bx)}). \\
    &\Lambda_{UCB\kappa}(\bx) = \mu(\bx) + \kappa \sigma(\bx).
\end{align*}

KG does not have a closed form. It is defined as the expected value of the posterior minimum:
\[
    \Lambda_{KG}(\bx) = \mathbb{E}_y[\mu^*(y | \bx)].
\]

Where $\mu^*(y | \bx)$ is the value of the the posterior mean having sampled $y$ at $\bx$. The distribution of $y$ is the posterior distribution of the GP.

\section{CONTROL VARIATES}
The general idea behind control variates is to find a covariate $g(y)$ with known mean and negative correlation with $f(y)$. The quantity $c(y) = f(y) + \beta g(y)$, known as a regression control variate (RCV), is estimated, and then de-biased afterwards.
If $\Var[f(y)]= \sigma_f$ and $\Var[g(y)]= \sigma_g$, then:
\[
\Var[c(y)] =  \sigma_f^2 + \beta^2\sigma_g^2 - 2\beta \Cov[f(y), g(y)] .
\]

The optimal value minimizing the variance of $c(y)$ is thus:
\[
\beta = \Cov[f(y), g(y)] / \sigma_h^2.
\]
In practice, both $ \Cov[f(y), g(y)]$ and $\sigma_h^2$ must be either estimated from samples of $f$ and $g$ or computed a-priori.

In the case of $k > 1$ control variates, we consider a vector of control variates $\bg(y) = [g_1(y), g_2, \dots, g_k(y)]^T$. Our estimator will have the form $c(y) = f(y) - \beta^T \bg(y)$, where $\beta$ is an length $k$ vector of constants. The optimal $\beta$ minimizing the variance of $c(y)$ is:
\[
\beta = \Sigma^{-1}_\bg * \sigma_{\bg, f},
\]
where $\Sigma_\bg$ is the covariance matrix of $\bg(y)$ and $\sigma_{\bg, f}$ is the vector of covariances between each variate and $f(y)$.

\begin{figure*}[t]
    \centering
    \includegraphics[width=\textwidth, trim={0cm 0cm 0cm 0cm}]{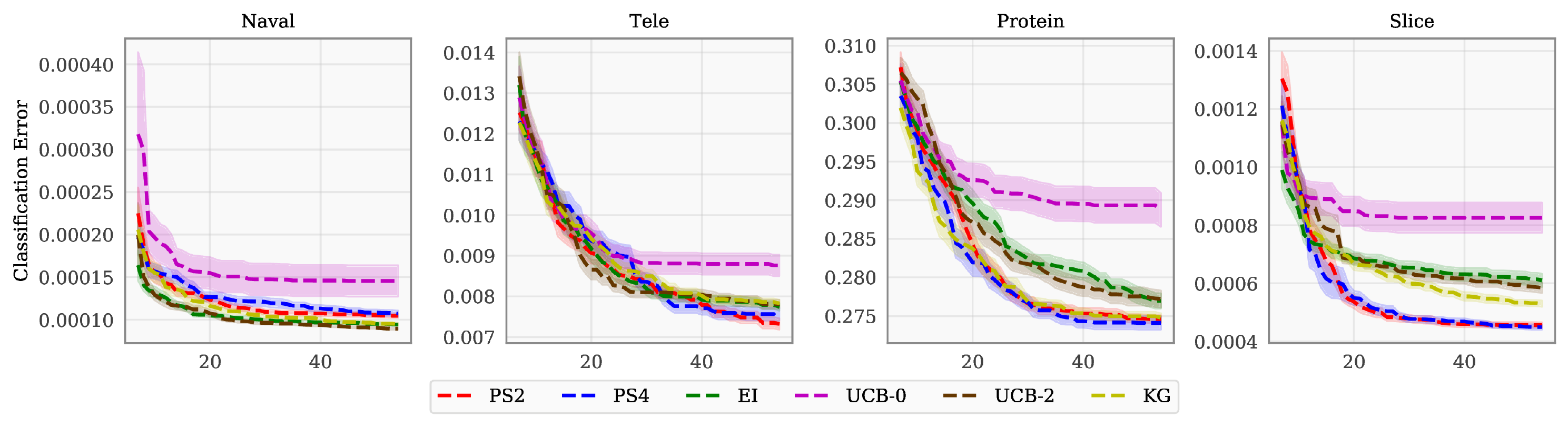}
    \caption{The classification error achieved by PS2 and PS4 is largely on par with, if not better than, the performance of EI, KG, and UCB variants. The only exception is the \textit{Tele} dataset. }\label{fig:nas_all}
\end{figure*}

\begin{figure*}[t]
    \centering
    \includegraphics[width=0.8\textwidth, trim={0cm 0cm 0cm 0cm}]{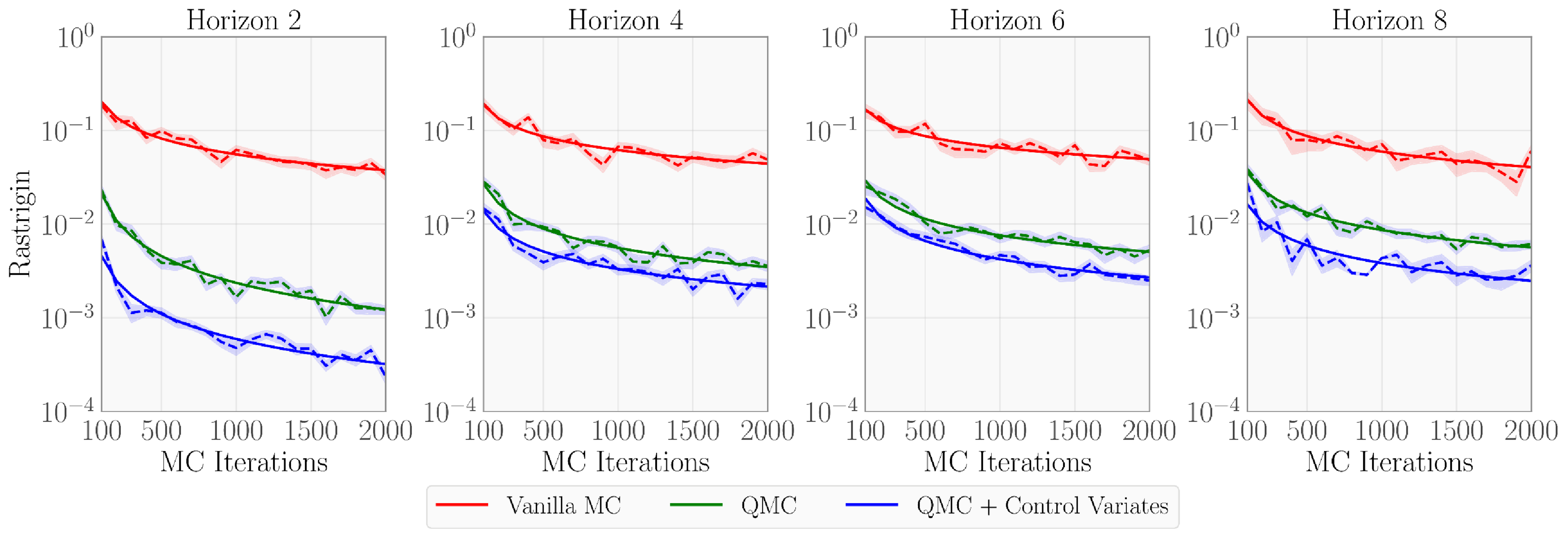}
    \caption{The estimation errors of MC (red), QMC (green), and QMC combined with control variates (blue).}\label{fig:ablation}
\end{figure*}

\section{NAS BENCHMARK}
The NAS benchmark is a tabular benchmark containing all possible hyperparameter configurations evaluated for a two-layer multi-layer perceptron on different datasets. The search space we consider is:
\begin{itemize}
    \item Batch size in  $\{8, 16, 32, 64\}$.
    \item Epochs in $\{ 10, 20, 30, 40, 50, 60, 70, 80, 90, 100 \}$.
    \item Layer 1 width in $\{16, 32, 64, 128, 256, 512\}$.
    \item Layer 2 width in $\{16, 32, 64, 128, 256, 512\}$.
\end{itemize}
The resulting search space is four-dimensional. We optimize over the unit hypercube $[0,1]^4$ and scale and round evaluation points to the corresponding NAS search space entry. Note that the NAS benchmark contains other hyperparameters as well, which we set to the default. These include the activation functions (default: tanh), the dropout (default: 0), the learning rate (default: 0.005), and the learning rate schedule (default: cosine).

The datasets in the NAS benchmark are all classification tasks taken from the UC Irvine repository for machine learning datasets. We run our method on all four in the NAS benchmark: \textit{Naval}, \textit{Tele}, \textit{Protein}, and \textit{Splice}. The achievable classification error for each dataset is different, so we compare methods by regret, which is defined as:

\[
\frac{y_{init} - y_{best}}{y_{init}},
\]

where $y_{init}$ is the starting value during optimization and $ y_{best}$ is the best observed value during iteration so far. For each dataset, we run BO using EI, KG, UCB0, UCB2, and our policy search methods for horizons 2 and 4, labeled PS2 and PS4 respectively. We replicate BO runs 50 times. In our main paper, we plotted the average regret among the four datasets, and PS2 and PS4 beat the competing methods. In this supplement, we plot the the individual classification errors for further clarity in Figure \ref{fig:nas_all}. We find the performance of both PS2 and PS4 performance are largely on par with, if not better than, the performance of EI, KG, and UCB variants.

\section{ABLATION STUDY}
Recall that we combine QMC and control variates to achieve high levels of variance reduction in the resulting Monte carlo estimator.

In Figure \ref{fig:ablation}, we empirically measure the individual impact of QMC and control variates. We roll out EI for horizons $2$, $4$, $6$, and $8$, and calculate the variance of estimators for MC sample sizes in $[100, 200, 300, \dots, 2000]$, using $50$ trials each.  We compare the Vanilla MC estimator, a QMC estimator, and a QMC estimator that also uses control variates. The underlying function is the Rastrigin function. As we mentioned before, the effectiveness of our control variates, which consist of myopic acquisition functions, are less effective as $h$ increases. As a whole, QMC contributes to a greater drop in variance.

\end{document}